\documentclass[conference]{IEEEtran}
\usepackage{graphicx}
\usepackage{hyperref} 
\usepackage{amsmath,amsthm,amssymb}
\usepackage{graphicx}
\usepackage{abstract}

\begin{document}

\title{The layer-wise L1 Loss Landscape of Neural Nets is more complex around local minima}

\author{\IEEEauthorblockN{Peter Hinz}\
\IEEEauthorblockA{ETH Zürich}
}

\maketitle

\begin{abstract}
	For fixed training data and network parameters in the other layers the L1 loss of a ReLU neural network as a function of the first layer's parameters is a piece-wise affine function. We use the Deep ReLU Simplex algorithm to iteratively minimize the loss monotonically on adjacent vertices and analyze the trajectory of these vertex positions. We empirically observe that in a neighbourhood around a local minimum, the iterations behave differently such that conclusions on loss level and proximity of the local minimum can be made before it has been found: Firstly the loss seems to decay exponentially slow at iterated adjacent vertices such that the loss level at the local minimum can be estimated from the loss levels of subsequently iterated vertices, and secondly we observe a strong increase of the vertex density around local minima. This could have far-reaching consequences for the design of new gradient-descent algorithms that might improve convergence rate by exploiting these facts.

\end{abstract}

\IEEEpeerreviewmaketitle

\section{INTRODUCTION AND MOTIVATION}
Due to its relevance for the design of efficient neural network training algorithms, a deeper understanding of the loss landscape of neural networks is of great interest and a topic of active research. However, being a non-convex function of a typically very high number of parameters which is usually explored only point-wise or path-wise by training algorithms, insight in its structure requires sophisticated ideas and techniques. 

Related literature covers both, theoretical results for specific settings and experimental studies. In \cite{pmlr-v70-nguyen17a} it is shown that for squared error loss, specific networks and limited training samples the training loss behaves nicely in the sense that almost all local minima are global.  The work \cite{li2017visualizing} studies loss visualization techniques and qualitatively relates geometrical properties, generalization error and trainability. The authors of \cite{im2017empirical} and \cite{yu2020Experimental} experimentally analyze the loss landscape, its properties such as curvature and Hessian matrix along the trajectories of various optimization algorithms and compare their resulting converged local minima. In this work we perform a novel analysis of the vertex structure of the layer-wise L1 loss of feed-forward ReLU neural networks.

Feed-forward ReLU neural networks with $L\in\mathbb{N}$ hidden layers and layer widths $(n_0,\ldots,n_{L+1})\in\mathbb{N}^{L+2}$ are functions $f_{\theta}:\mathbb{R}^{n_0}\to\mathbb{R}^{n_{L+1}}$ of the form 
\begin{equation}
	\label{eq:f}
	f_{\theta}(x)=W_{L+1}g_{\theta_L}\circ\cdots\circ g_{\theta_1}(x)+b_{L+1}\quad \textnormal{ for }x\in\mathbb{R}^{n_0}
\end{equation} 
with layer transition functions $g_{\theta_l}:\mathbb{R}^{n_{l-1}}\to\mathbb{R}^{n_l}, x\mapsto \textnormal{ReLU}(W_lx+b_l)$, $l\in\left\{ 1,\cdots,L \right\}$ that use a coordinate-wise application of the activation function $\textnormal{ReLU}:t\mapsto \max(0,t)$. The parameter $\theta$ determines the weight and bias matrices, i.e. $\theta=(\theta_1,\ldots,\theta_{L+1})$ with $\theta_l=(W_l,b_l)\in\mathbb{R}^{n_{l}\times n_{l-1}}\times \mathbb{R}^{n_{l}}$, $l\in\left\{ 1,\ldots,L +1 \right\}$. These functions are piece-wise affine on convex regions which have the form of a feasible region of linear inequalities.

\begin{figure}[htpb]
\centering
\includegraphics[width=0.8\linewidth]{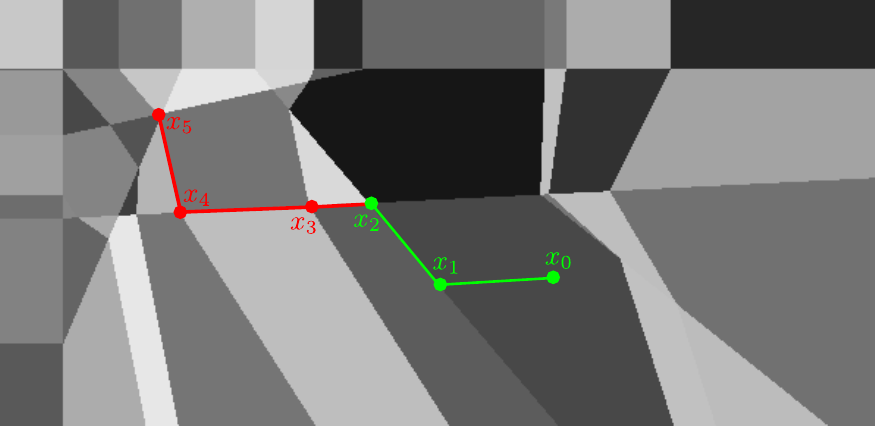}
\caption{Qualitative illustration of the DRLSimplex algorithm iterating on vertices of a continuous, piece-wise affine function given in the form of a feed-forward ReLU neural network. In the first phase (green) a vertex is found similarly to standard linear programming and in subsequent iterations then the feasible regions is changed.}
\label{fig:DRLSimplex}
\end{figure}

We recently published the \emph{Deep ReLU Simplex (DRLSimplex)} method \cite{deepReLUProgramming} which is a novel algorithm for the iterative minimization of such functions. It iterates on vertices of feed-forward ReLU neural networks and is closely related to the simplex algorithm from linear programming. The main extension of the DRLSimplex over the standard simplex algorithm is the ability to efficiently change the feasible region at a vertex. The path of visited vertices has the important property that the corresponding function values are non-increasing. Figure~\ref{fig:DRLSimplex} qualitatively depicts the iteration behaviour: In the initial phase, a set of active constraints is constructed while the initial affine region determined by the starting position is not left and takes the role of the feasible region in linear programming (steps depicted in green). After a vertex has been found, the feasible region is changed to some of the adjacent regions and the iteration continues by finding an adjacent vertex and changing the feasible region (steps depicted in red). The next vertex is chosen such that the objective function is decreased and the iteration terminates when the objective function value is larger at all adjacent vertices. This way, after a finite number of iterations the DRLSimplex algorithm converges at a local minimum.

The L1 training loss 
\begin{equation}
	\label{eq:l1TrainingLoss}
	L(\theta)=  \sum_{i=1}^N\sum_{j=1}^{n_{L+1}}|(y_i)_j-f_\theta(x_i)_j|
\end{equation}
of $N\in\mathbb{N}$ samples $(x_1,y_1),\ldots,(x_N,y_N)\in\mathbb{R}^{n_0}\times \mathbb{R}^{n_{L+1}}$ itself can be written as a ReLU feed-forward neural network when only one layer shall be optimized while keeping the other layer parameters fixed. Therefore, we can apply the DRLSimplex algorithm to generate a sequence of network parameters (weight matrix and bias vector entries) corresponding to a specific layer which have non-increasing L1 loss value. We analyze the behaviour of this sequence of vertex positions and loss values as we approximate a local minimum. The DRLSimplex algorithm is well suited for this kind of analysis due to the following reasons:
\begin{enumerate}
	\item Compared to gradient descent, where the step size is usually a function of past step sizes or decaying at a predetermined rate, the step size of the DRLSimplex algorithm is automatically determined in every iteration by the proximity of the next vertex in the selected direction. This way we directly get access to local properties of the underlying objective function, i.e. the loss landscape.
	\item Gradient descent algorithms do not guarantee that the training loss values decay in every iteration. Without proper step size control, they do not necessarily converge. In contrast, the DRLSimplex algorithm is guaranteed to terminate in a local minimum without any adjustments. 
\end{enumerate}
From these two facts it is clear that the DRLSimplex algorithm provides a finite loss-decreasing trajectory of adjacent vertices which has a direct relation to the converged local minimum and is \emph{not} affected by an arbitrary externally defined factors such as step size control rules. This makes this algorithm perfectly suited for an analysis on how local vertex density and corresponding quantities behave as we approximate local minima.

\section{SIMULATION AND RESULTS}

For simplicity we only consider the case of first layer L1 optimization, i.e. we search a local minimum of the function
\begin{equation}
	\label{eq:layertrainingloss}
	L_{1,\theta^*}:\theta_1\mapsto L\left((\theta_1,\theta^*_2,\ldots,\theta^*_{L+1})  \right).
\end{equation} with fixed weight matrices and bias vectors $\theta_2^*,\ldots,\theta^*_{L+1}$ for the other layers. Note that this is not a strong restriction since by equations~\eqref{eq:l1TrainingLoss} and \eqref{eq:f} for every layer the L1 loss optimization of this layer's parameters can be interpreted as first-layer parameter training of the smaller network starting with that layer and using different training predictors transformed by the lower layers  instead of the original predictors.

As we have shown in \cite{deepReLUProgramming} our DRLSimplex can be implemented efficiently by making use of the insight and best-practices concerning performance and numerical stability that evolved over decades in the field of linear programming. However, such an implementation still has to be written and tested such that at the present state we rely on complete matrix inversion instead of simplex tableau modifications to favour correctness over performance. Therefore we were are only able perform our analysis on a small scale.

Our simulation was carried out as follows: We picked the network architecture $L=4$ and $(n_0,\ldots,n_5)=(4,5,4,3,2,1)$ such that there are $n_1(n_0+1)=25$ first layer parameters consisting of the $5\times 4$ weight matrix and the $5$-element bias vector that are applied to the network input. For the fixed other layers' parameters $\theta^*_2,\ldots,\theta^*_{L+1}$ we sampled all entries from a uniform distribution on $[-1,1]$. We then generated $N=500$ random training samples $(x_1,y_1),\ldots,(x_{500},y_{500})\in\mathbb{R}^4\times \mathbb{R}$ all with entries from independent uniform distributions on $[-3,3]$. Finally we sampled a random initial position with uniformly distributed entries on $[-20,20]$ in the $25$-dimensional parameter space of first layer parameters to start the iteration there using our DRLSimplex algorithm. We performed our analysis several times also with a different number of layers and other layer widths and the results were always similar. We will illustrate our findings in a representative simulated case of such a optimization trajectory of vertices ending in a local minimum $\hat\theta_1$ generated by our DRLSimplex algorithm.

\begin{figure}[htpb]
\centering
\includegraphics[width=0.8\linewidth]{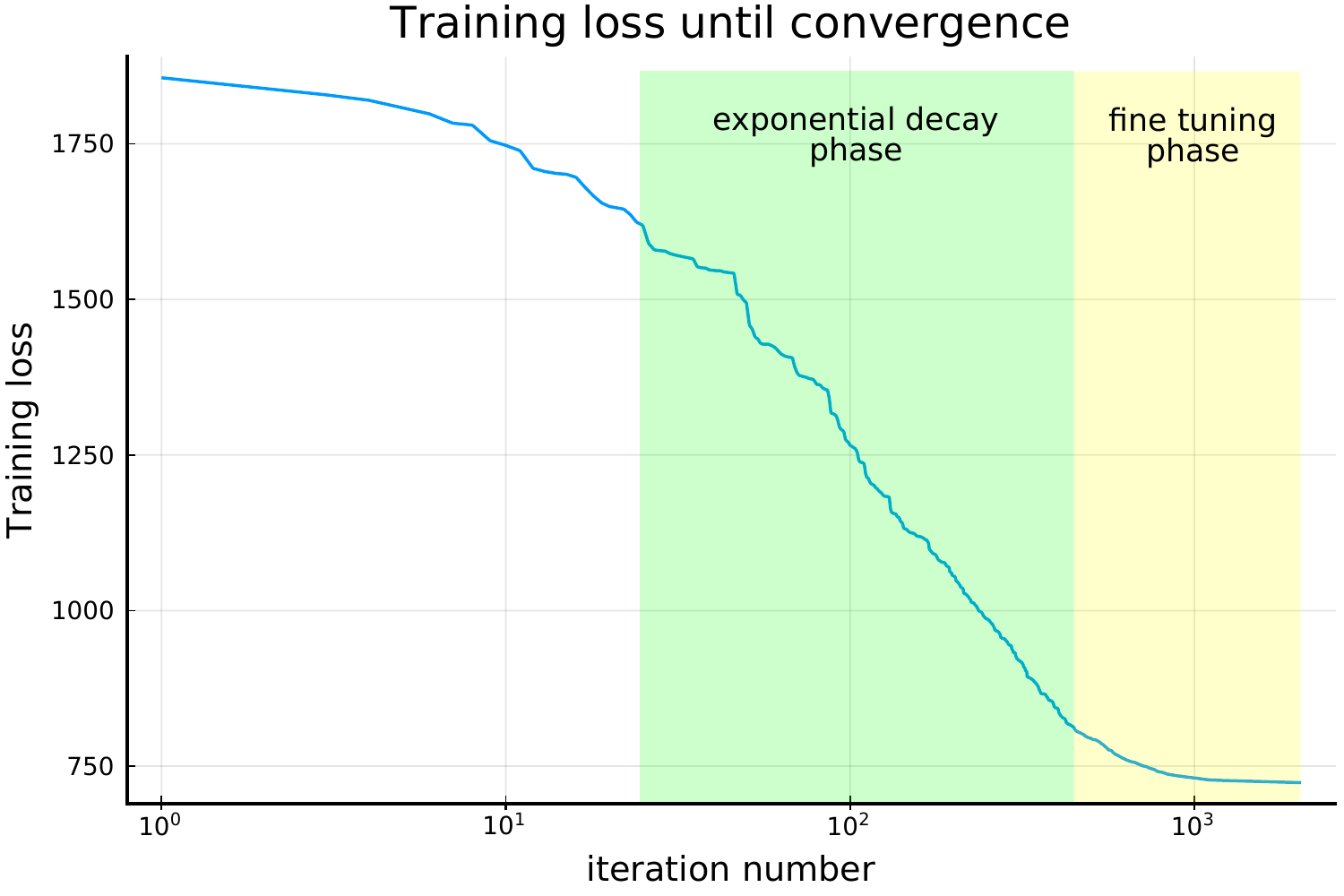}
\caption{First layer training loss \eqref{eq:layertrainingloss} at iterated first layer parameters returned by our DRLSimplex algorithm against iteration number in logarithmic scale. The loss sequence is non-increasing by construction.}
\label{fig:loss}
\end{figure}

\begin{figure}[htpb]
\centering
\includegraphics[width=0.8\linewidth]{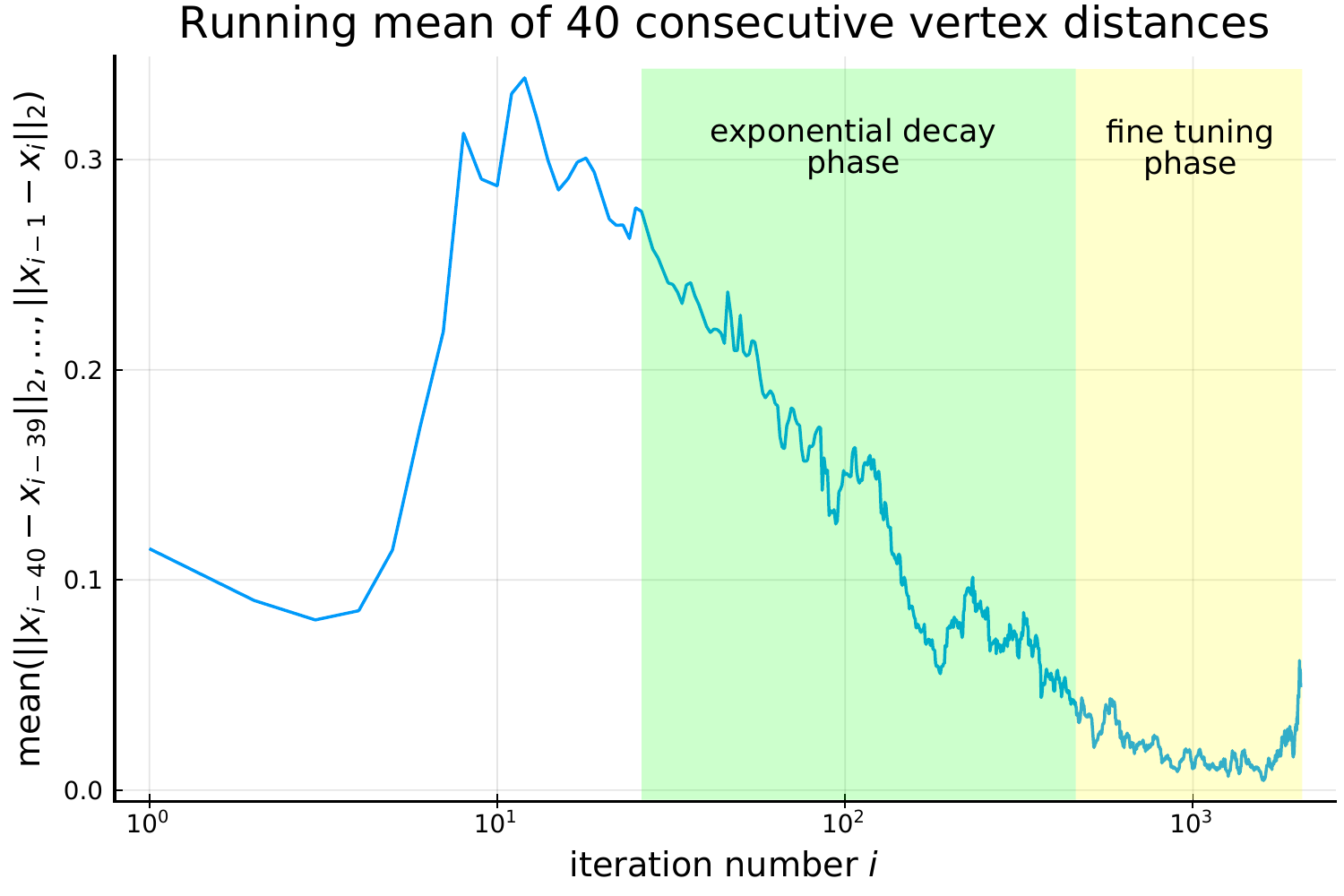}
\caption{Running mean of 40 consecutive L2-distances of iterated first layer parameters (the first 39 mean values are based on fewer observations). }
\label{fig:consecutivedist}
\end{figure}

\begin{figure}[htpb]
\centering
\includegraphics[width=0.8\linewidth]{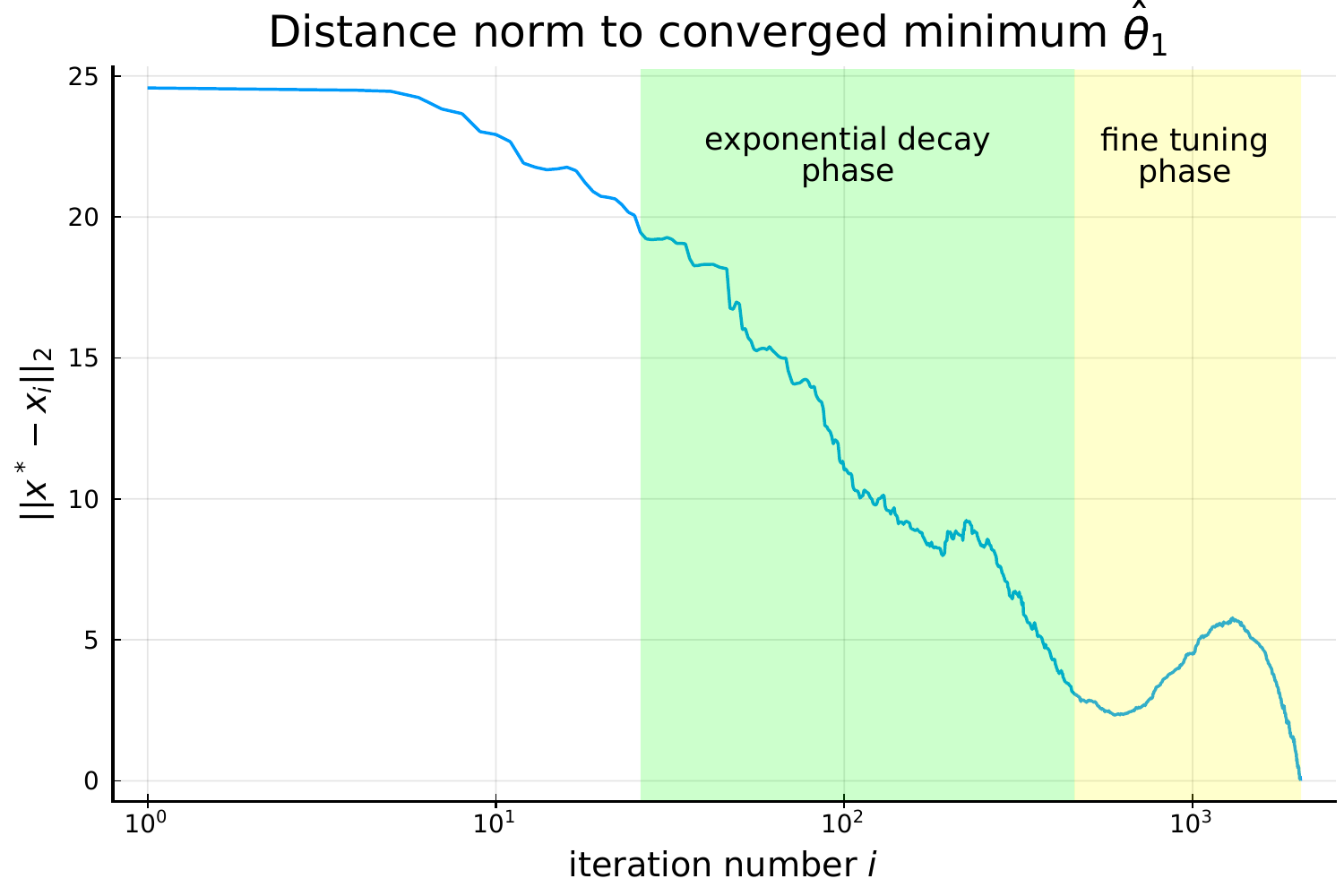}
\caption{L2-distance of iterated first-layer parameters to the final converged local minimum $\hat\theta_1$ (last element of sequence of iterated adjacent vertices) returned by the DRLSimplex algorithm.}
\label{fig:distnorm}
\end{figure}

In Figures~\ref{fig:loss}, \ref{fig:consecutivedist} and \ref{fig:distnorm} we depict the path of the training loss, a smoothed curve of the L2 distance of consecutive iterated parameter estimates and the distance to the local minimum vertex that our algorithm has finally converged to. At the beginning of the training process, our DRLSimplex algorithm iterates in the boundary of the same initial feasible region while successively adding hyperplane constraints until a vertex has been found. This process generates the first 25 iteration positions. In consequence, all iterated parameter estimates starting with index 25 are vertices of the layer-wise loss~\eqref{eq:layertrainingloss}. 

After a number of initial training iterations we observe that the remaining training process can be partitioned in two phases:
\begin{enumerate}
	\item In the \emph{exponential decay phase} we observe that the training loss, the L2 distance between successive iterated parameter estimates and the distance to the finally converged minimum all decrease roughly exponentially against the iteration index as can be seen in Figures~\ref{fig:loss}, \ref{fig:consecutivedist} and \ref{fig:distnorm} respectively. This is a typical observation we made in each of our repeated simulations. In this phase we approximate the local minimum in a systematic and stable way. The fact that our considered iterations are adjacent vertices of $L_{1,\theta^*}$ allows us to conclude that the vertex density increases exponentially during this phase as we approximate the local minimum $\hat \theta_1$.
	\item In the \emph{fine tuning phase} we observe a slower loss decrease than in the exponential decay tuning phase. The step size of consecutive vertices iterations and the distance to the final local minimum do not show a systematic behaviour across our simulations. We conclude that in contrast to the exponential decay phase, this phase is highly individual and case-specific.
\end{enumerate}

\section{INTERPRETATION}
The decreasing distance observed in the trajectory of adjacent vertices approaching a local minimum corresponds to a higher vertex density, i.e. more and smaller affine regions around local minima. If the number of regions is considered as a natural measure of function complexity, this observation indicates that the function complexity of the considered layer-wise L1 loss function is greater around local minima.

Furthermore, the two phases we observed in the training path of iterated parameter estimates indicate that the neighbourhood of local minima of the layer-wise L1 loss is also two fold as shown in Figure~\ref{fig:neighbourhood}:
\begin{figure}[htpb]
\centering
\includegraphics[width=0.8\linewidth]{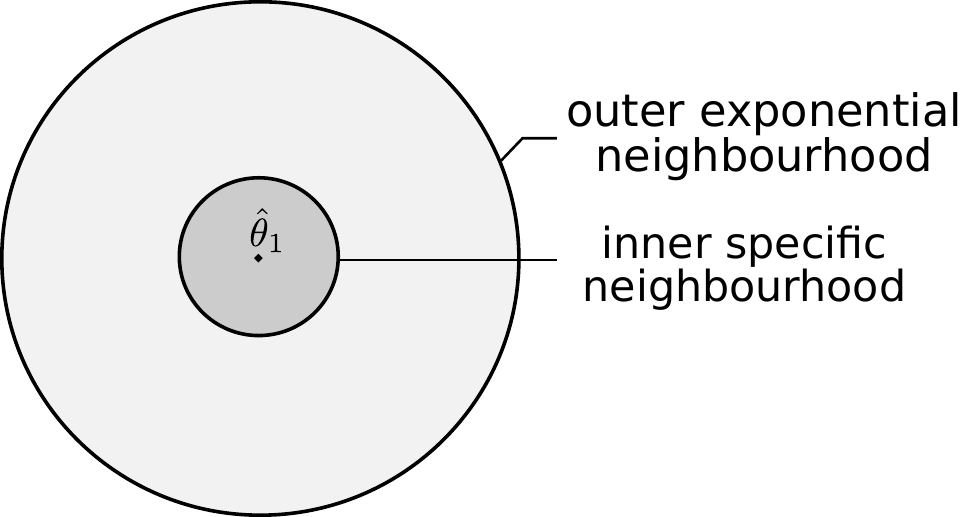}
\caption{Schematic interpretation of neighbourhood of layer-wise L1 loss local minima $\hat\theta_1$ based on our results.}
\label{fig:neighbourhood}
\end{figure}
\begin{enumerate}
	\item There is an \emph{outer exponential neighbourhood} that behaves essentially similarly for all local minima we observed in the sense that the vertex distance, the distance to the local minimum and the loss level decay exponentially on a sequence of adjacent vertices approaching the local minimum. 
	\item Furthermore there is an \emph{inner specific neighbourhood} where the loss decay of such a sequence is slower and both, the vertex distance and the distance to the local minimum are individual and do not have a typical behaviour.
\end{enumerate}

The general behaviour repeatedly observed in the outer exponential neighbourhood suggests that the vertex structure of most local minima behaves consistently in a similar way in their neighbourhood. This fact could be used to extract information about a local minimum based on the iterated parameter values. For example, we can already give a good estimate on the loss value at a local minimum $\hat\theta_1$ that the DRLSimplex algorithm finally converges to before that local minimum has been found, based on an exponential decay fit as soon as this decay regime is recognized. In other words, without having found a local minimum the trajectory of vertices returned by our DRLSimplex algorithm already reveals insight in its loss level. 

In future research one could analyze this \emph{information leak} of local minima onto their neighbourhood. Furthermore, in the light of neural network training one can speculate on the relation of over-fitting and the inner specific neighbourhood, and a deeper analysis would be a second interesting research direction.

\section{CONCLUSION}
In this work we considered the L1 training loss of ReLU feed-forward neural network as a function of the first layer's parameters and for fixed parameters in the other layers. This function is piece-wise affine on convex regions and induces a vertex structure. We used the DRLSimplex algorithm to obtain a loss-decreasing sequence of adjacent vertices ending in a local minimum. Our numerical experiments results suggest that locally around local minima the vertex density of the layer-wise L1 loss is larger, which indicates that there are more and smaller regions located around local minima and this can be interpreted as higher function complexity.

A further insight is the fact that that layer-wise L1 loss local minima seem to leak information to their neighbourhood and that this information can be retrieved from the steps of an iterating optimization procedure. We used our DRLSimplex algorithm to get access to local properties of the loss function such as vertex density and in future research it might be interesting to analyze to what extent also the trajectory of iteration parameters generated by popular training procedures carries information on proximity, direction or loss level of nearby local minima that could be used to adjust step size control in gradient-descent like minimization procedures and ultimately improve convergence rate of neural network training algorithms.

\bibliographystyle{ieeetr}
\bibliography{../bib/all} 

\begin{thebibliography}{1}

\bibitem{pmlr-v70-nguyen17a}
Q.~Nguyen and M.~Hein, ``{The Loss Surface of Deep and Wide Neural Networks},''
  vol.~70, pp.~2603--2612, 06--11 Aug 2017.

\bibitem{li2017visualizing}
H.~Li, Z.~Xu, G.~Taylor, C.~Studer, and T.~Goldstein, ``{Visualizing the loss
  landscape of neural nets},'' {\em arXiv preprint arXiv:1712.09913}, 2017.

\bibitem{im2017empirical}
D.~J. Im, M.~Tao, and K.~Branson, ``{An empirical analysis of the optimization
  of deep network loss surfaces},'' {\em arXiv preprint arXiv:1612.04010},
  2017.

\bibitem{yu2020Experimental}
Q.~Yuan and N.~Xiao, ``{Experimental exploration on loss surface of deep neural
  network},'' {\em International Journal of Imaging Systems and Technology},
  vol.~30, no.~4, pp.~860--873, 2020.

\bibitem{deepReLUProgramming}
P.~Hinz and S.~van~de Geer, ``{Deep ReLU Programming},'' {\em arXiv preprint
  arXiv:2011.14895}, 2021.

\end{thebibliography}
\end{document}